\documentclass[10pt,journal,compsoc]{IEEEtran}

\ifCLASSOPTIONcompsoc
  \usepackage{cite}
\else
  \usepackage{cite}
\fi

\ifCLASSINFOpdf
\else
\fi

\usepackage{amsmath,amssymb,amsfonts}
\usepackage{algorithmic}
\usepackage{graphicx}
\usepackage{textcomp}

\usepackage{balance}
\usepackage[flushleft]{threeparttable}
\usepackage{multirow}
\usepackage{booktabs}
\usepackage{nth}

\def\BibTeX{{\rm B\kern-.05em{\sc i\kern-.025em b}\kern-.08em
    T\kern-.1667em\lower.7ex\hbox{E}\kern-.125emX}}

\DeclareMathOperator*{\argmin}{\arg\min}
\DeclareMathOperator*{\argmax}{\arg\max}

\begin{document}
%
\title{FlowSAN: Privacy-enhancing Semi-Adversarial Networks to Confound Arbitrary Face-based Gender Classifiers}

\author{Vahid Mirjalili,
        Sebastian Raschka,
        and~Arun Ross
\IEEEcompsocitemizethanks{\IEEEcompsocthanksitem V. Mirjalili and A. Ross are with the Department
of Computer Science and Engineering, Michigan State University, East Lansing,
MI, 48824.\protect\\
E-mail: \{mirjalil,rossarun\}@cse.msu.edu
\IEEEcompsocthanksitem {S. Raschka is with the Department of Statistics, University of Wisconsin -- Madison, Madison, Wisconsin. \protect\\
E-mail: srachka@wisc.edu}}
}

\markboth{}{}%
\IEEEtitleabstractindextext{%
\begin{abstract}
Privacy concerns in the modern digital age have prompted researchers to develop techniques that allow users to selectively suppress certain information in collected data  while allowing for other information to be extracted. In this regard, Semi-Adversarial Networks (SAN) have recently emerged as a method for imparting soft-biometric privacy to face images. SAN enables modifications of input face images so that the resulting face images can still be reliably used by arbitrary conventional face matchers for recognition purposes, while attribute classifiers, such as gender classifiers, are confounded. However, the generalizability of SANs across \emph{arbitrary} gender classifiers has remained an open concern. 
In this work, we propose a new method, FlowSAN, for allowing SANs to generalize to multiple unseen gender classifiers.
We propose combining a diverse set of  SAN models to compensate each other's weaknesses, thereby, forming a robust model with improved generalization capability. 
Extensive experiments using different unseen gender classifiers and face matchers demonstrate the efficacy of the proposed paradigm in imparting gender privacy to face images.
\end{abstract}

\begin{IEEEkeywords}
Biometrics, Face Image, Semi-Adversarial Networks, SAN, Gender, Privacy, Adversarial, Deep Learning.
\end{IEEEkeywords}}

\maketitle

\IEEEdisplaynontitleabstractindextext

%
\IEEEpeerreviewmaketitle

\IEEEraisesectionheading{\section{Introduction}\label{sec:introduction}}

%
%
%
%

\IEEEPARstart{F}{ace} images of individuals contain valuable information unique to themselves that facilitates biometric face recognition. In addition, other auxiliary information such as age, gender, and race, which are called {\em soft-biometrics}, can also be extracted from face images using machine learning techniques~\cite{jain_introduction_2011,dantcheva_what_2016,sundararajan_deep_2018}.  Face recognition involves comparing features extracted from a pair of face images, using a {\em face matcher}, to determine their degree of similarity~\cite{jain_introduction_2011,chang_face_2003}.  The increasing use of face recognition in various applications has brought the issue of data privacy to the forefront~\cite{jain_biometrics:_2006,ratha_enhancing_2001,natgunanathan_protection_2016_long,morales_sensitivenets_2019,chhabra_anonymizing_2018,sim_controllable_2015,medcn_selective_2018,meden_ksamenet_2018,terhorst_unsupervised_2019,wu_privacy_2019,zheng_survey_2018,li_method_2018,yang_using_2018,jia_right_2018}. 
While extracting soft-biometric information can be useful in many applications~\cite{swearingen_label_2017}, we should note that such information can be abused in several ways, such as profiling users, targeted advertisement, and increasing the risk of linkage attacks~\cite{acquisti_predicting_2009}. Furthermore, extracting this information without the users' consent may be viewed as a violation of their privacy.
One aspect of privacy involves granting users the right to determine which personal information to reveal and which to conceal~\cite{kindt_privacy_2016,acquisti_what_2013}. In this regard, \emph{soft-biometric privacy} was introduced as a means for preserving the biometric utility of face images, while confounding soft-biometric information, such as gender characteristics~\cite{othman_privacy_2014,mirjalili_soft_2017}.

Recently, European Union's General Data Protection Regulation
(GDPR)~\cite{eu_gdpr_2016} has come to effect. One of its goals is to protect the data collected from European users and to regulate its usage. To this effect, it enforces any entity (individual or group) collecting data from European users to disclose the type-of-data collected, the intended usage, and the data-processing techniques that will be used. 
Accordingly, GDPR prohibits any processing of individuals' information beyond the {\em stated purpose} at the time of data collection.
For example, consider a scenario where users of an application or service can optionally withhold their gender information; however, such information could still be  extracted automatically from their biometric data~\cite{bobeldyk_predicting_2018,mansanet_local_2016,jia_gender_2016,lyle_soft_2010,arora_robust_2018,chhabra_data_2018,bobeldyk_predicting_2019,sgroi_prediction_2013,nixon_soft_2015}.

In the context of GDPR, biometric data of individuals, such as face photos or fingerprints, are collected solely for the purpose of user recognition, without acquiring other demographic information such as age, gender, and ethnicity. 
In such a scenario, applying data processing techniques that allow extracting such sensitive information automatically from a person's biometric data~\cite{jain_introduction_2011,dantcheva_what_2016,du_garp_2014,sun_demographic_2017_long,bobeldyk_predicting_2019,bobeldyk_analyzing_2019,lagree_predicting_2011,badawi_fingerprint_2006,raschka_python_2017} without their knowledge and consent is a violation of the users' privacy. 
While GDPR prohibits  unsolicited data extraction from European users, the possibility of unlawful data collection still remains and can ultimately lead to negative societal, economic, and political consequences~\cite{facebook_scandal_2018,garvie_perpetual_2016,alvi_turning_2018}.

Previously, we developed Semi-Adversarial Networks (SAN)~\cite{mirjalili_semi_2018} for imparting soft-biometric privacy to face images, where a face image is modified such that the matching utility of the modified face image is retained while the automatic extraction of gender information is confounded. 
In our previous work~\cite{mirjalili_semi_2018}, we empirically showed that the ability to predict gender information, using an unseen gender classifier from outputs of the SAN model, is successfully diminished.
In~\cite{mirjalili_gender_2018}, we defined the generalizability of the SAN model as its ability to confound arbitrary unseen\footnote{The term ``unseen'' indicates that a certain classifier (or face matcher) was not used during the training stage. On the contrary, the term ``auxiliary'' in this paper refers to the classifier (or face matcher) that is used during the training phase.} gender classifiers. Generalizability is an important property for real-world privacy applications since the lack thereof implies that there exists at least one gender classifier that can still reliably estimate the gender attribute from outputs of the SAN model and, therefore, jeopardizes the privacy of users.
In order to address the generalizability issue of SAN models, in this paper, we propose the FlowSAN model,  
that progressively degrades the performance of unseen gender classifiers. 
Extensive experiments on a variety of independent gender classifiers and face image datasets show that the proposed FlowSAN method (Fig.~\ref{fig:stacking-san-idea}) results in a substantially improved generalization performance compared to the original SAN method with regard to concealing gender information while retaining face matching utility.


\section{Related Work}
With regard to privacy concerns in recent years, a new line of research has emerged that focuses on methods for imparting soft-biometric privacy to biometric data and face images in particular~\cite{othman_privacy_2014,mirjalili_soft_2017,chhabra_anonymizing_2018,morales_sensitivenets_2019,sim_controllable_2015,proteek_mitigating_2019}. Othman and Ross~\cite{othman_privacy_2014} first proposed an approach for mixing input face images with candidate images of the opposite gender using Active Shape Model~\cite{cootes_active_1995}. 
Subsequently, Mirjalili and Ross~\cite{mirjalili_soft_2017} developed a scheme that modifies an input face image using adversarial perturbations~\cite{rozsa_facial_2016} where the performance of a given gender classifier was confounded while the performance of a face matcher was retained. 
Chhabra~et~al.~\cite{chhabra_anonymizing_2018} later extended this research by including multiple attribute classifiers. They applied additive perturbations to face images to either preserve or suppress certain soft-biometric attributes~\cite{chhabra_anonymizing_2018}. 
While these proposed schemes successfully confound a target attribute classifier, they fail to generalize to unseen attribute classifiers. Thus,  soft-biometric attributes remain susceptible to extraction by unseen classifiers. 

In order to derive perturbations that are transferable to unseen gender classifiers, Mirjalili~et~al.~\cite{mirjalili_semi_2018} designed a convolutional autoencoder that modifies input face images such that an auxiliary face matcher still retains good matching performance on the modified output image while confounding an auxiliary gender classifier. As a result, since the output of their model is adversarial to one classifier and not to the other, the architecture is referred to as Semi-Adversarial Networks (SAN). The SAN model was shown to be able to derive perturbations that are transferable to two unseen gender classifiers. 
In~\cite{mirjalili_gender_2018}, we investigated the generalizability of SAN models across multiple arbitrary gender classifiers and formulated an ensemble SAN model with a training scheme based on different data augmentation techniques, to enhance diversity in the ensemble of SAN models.  Furthermore, we explored the effectiveness of randomly selecting a perturbed image from an ensemble of SAN models, which we refer to as Ens-Gibbs~\cite{mirjalili_gender_2018}.

While these methods directly apply perturbations to face images, recently, new techniques have emerged where  perturbations were applied to face representation vectors computed by face matchers~\cite{morales_sensitivenets_2019,terhorst_unsupervised_2019}. 
In particular, Morales et al.~\cite{morales_sensitivenets_2019} proposed a neural-network-based model, called SensitiveNet, that is able to remove soft-biometric information from face representation vectors. Therefore, any attribute classifier trained on face representation vectors may not be able to extract such sensitive information. However, these methods are based on the assumption that only face representation vectors are stored in a biometric database. This scheme is not desirable in many applications since only storing face representations results in 1) losing human interpretability, and 2) losing  backward matching compatibility when the face matcher is updated. An overview of existing techniques and their properties (transferability, generalization to arbitrary attribute classifiers, and retaining matching utility) is shown in Table~\ref{tab:relatedwork}.

\begin{figure}
\begin{center}
   \includegraphics[width=1\linewidth]{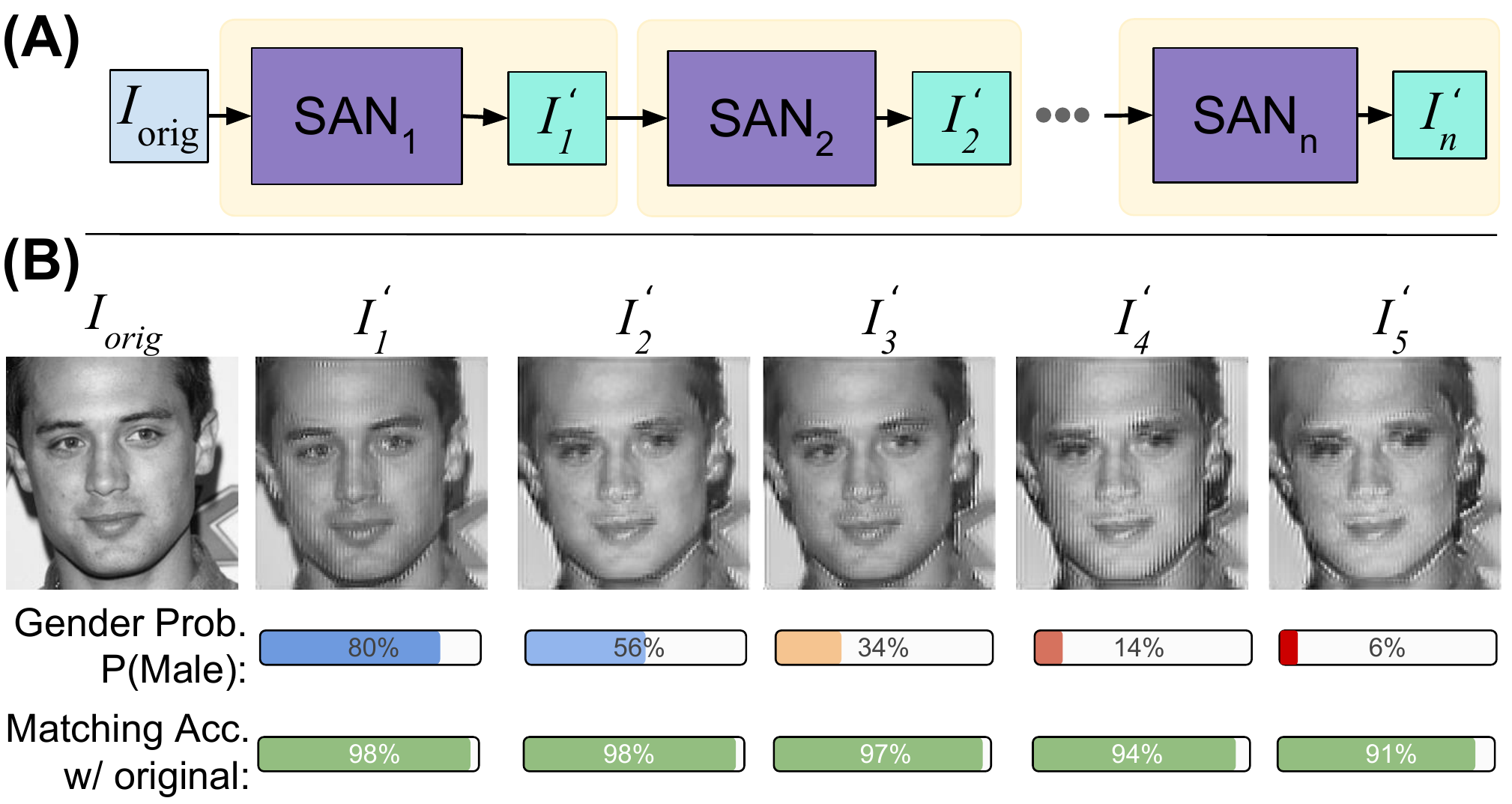}
\end{center}
   \caption{Illustration of the FlowSAN model, which sequentially combines individual SAN models in order to sequentially perturb a previously unseen gender classifier, while the performance of an unseen face matcher is preserved. A: An input gray-scale face image $I_{\text{orig}}$ is passed to the first SAN model ($\text{SAN}_1$) in the ensemble. The output image of $\text{SAN}_1$, $I'_1$, is then passed to the second SAN model in the ensemble, $\text{SAN}_2$, and so forth. B: An unmodified face image from the CelebA~\cite{liu_deep_2015_long} dataset ($I_\text{orig}$) and the perturbed variants $I'_i$ after passing it through the different SAN models sequentially. The gender prediction results measured as probability of being male ($P(\text{Male})$) as well as the face match score between the original ($I_{\text{orig}}$) and the perturbed images ($I'_{i}$) are shown.}
\label{fig:stacking-san-idea}
\end{figure}

\begin{table*}[h]
\centering
\caption{\label{tab:relatedwork}Overview of existing methods for imparting soft-biometric privacy and their comparison based on three criteria: transferability, generalizability, and retention of matching performance; transferability refers to the ability to generate perturbations that can successfully confound a different gender classifier, whereas generalizability is a stronger criterion for the ability to confound \emph{any} arbitrary unseen gender classifier.}
\centering
\begin{tabular}{l||ccccc}\hline
\textbf{Authors} & \textbf{Domain} &\textbf{Proposed Method} & \textbf{Transferable} & \textbf{Generalizable} & \textbf{Matching Performance} \\\hline
Othman and Ross~\cite{othman_privacy_2014} & Face images & Mixing faces of opposite gender & Yes & Yes & Severely degraded \\ 
Sim and Li~\cite{sim_controllable_2015}& Face images & Multimodal Discriminant Analysis & Yes & Yes & Severely degraded\\
Mirjalili et al.~\cite{mirjalili_soft_2017}& Face images  & Adversarial perturbations & No & No & Mostly retained \\ 
Mirjalili et al.~\cite{mirjalili_semi_2018}& Face images  & Semi-Adversarial Networks & Yes & No & Mostly retained \\
Chhabra et al.~\cite{chhabra_anonymizing_2018} & Face images & Adversarial perturbations &  No & No & Mostly retained \\
Mirjalili et al.~\cite{mirjalili_gender_2018} & Face images & Ensemble of SAN models & Yes &  Yes & Mostly retained \\ 
Morales et al.~\cite{morales_sensitivenets_2019} & Face representations & SensitiveNet & Yes & Yes & Mostly retained \\ 
Terh{\"o}rst et al.~\cite{terhorst_unsupervised_2019} & Face representations & Noise transformation & Yes & Yes & Mostly retained \\ \hline
\end{tabular}
\end{table*}

In this work, we address the generalization issue of the SAN method using a novel stacking paradigm that will successively enhance the perturbations for confounding an arbitrary unseen gender classifier as illustrated in Fig.~\ref{fig:stacking-san-idea}. We refer to this method as FlowSAN. 
The primary contributions of this work are as follows:
\begin{itemize}
\item Designing the FlowSAN model that can successively degrade the performance of arbitrary unseen gender classifiers;
\item Generalizing the FlowSAN model to multiple arbitrary gender classifiers;
\item Demonstrating the practicality and efficacy of the proposed approach in confounding the gender information for real-world privacy applications via extensive experiments involving broad and diverse sets of datasets.
\end{itemize}

\section{Proposed Method}

\begin{figure}
\begin{center}
   \includegraphics[width=0.96\linewidth]{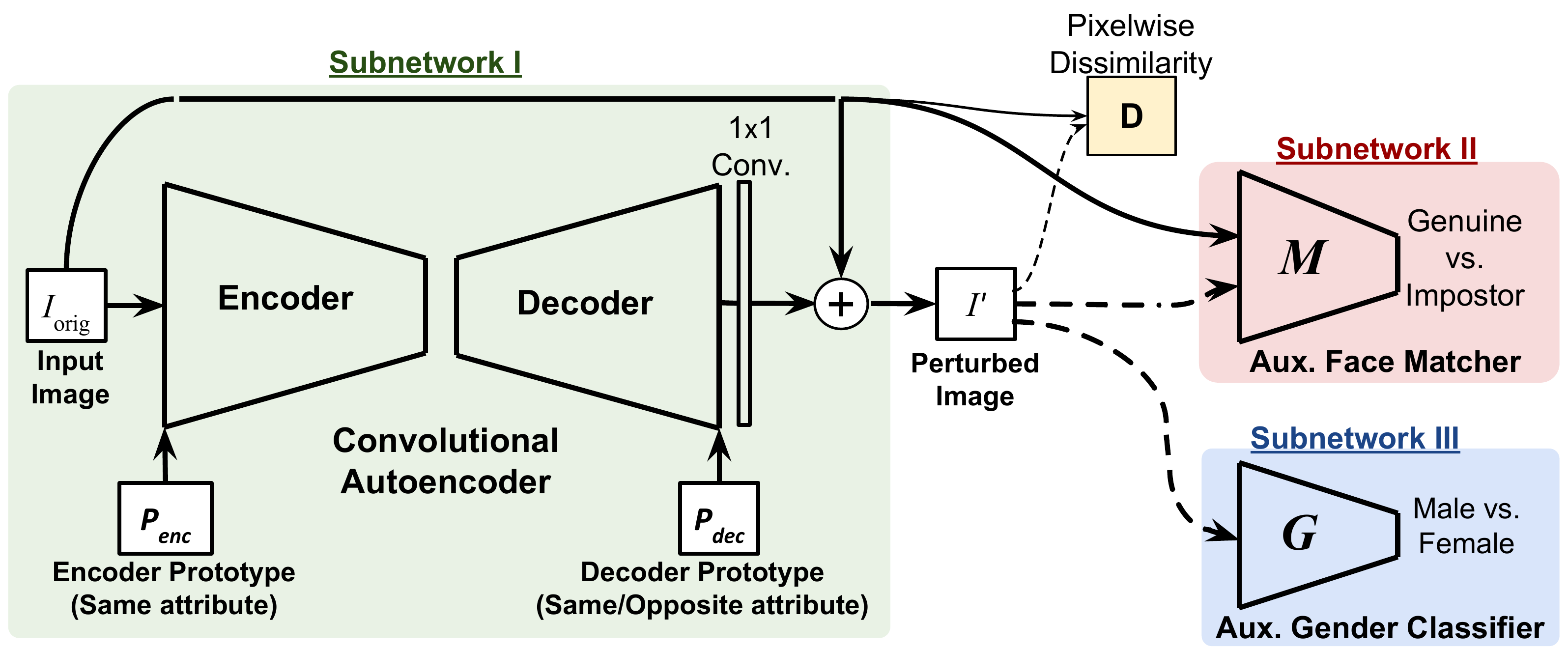}
\end{center}
   \caption{Architecture of the original SAN model~\cite{mirjalili_semi_2018} composed of three subnetworks: I: a convolutional autoencoder~\cite{bengio_learning_2009}, II: an auxiliary face matcher ($M$), and III: an auxiliary gender classifier ($G$). In addition, the unit $D$ computes the pixelwise dissimilarity between input and perturbed images during model training.}
\label{fig:san-arch}
\end{figure}

{\bf Original SAN model}~\cite{mirjalili_semi_2018}:  The SAN model for imparting gender privacy to face images was first proposed in~\cite{mirjalili_semi_2018}, and the overall architecture is shown in Fig.~\ref{fig:san-arch}. The SAN model leverages pre-computed {\bf face prototypes}, which are average face images for each gender. SAN consists of three subnetworks: 1) a {\bf convolutional autoencoder} that perturbs an input face image via face prototypes, 2) an {\bf auxiliary face matcher}, which is a convolutional neural network (CNN), and 3) a CNN-based {\bf auxiliary gender classifier}.  The input to the convolutional autoencoder is a gray-scale\footnote{Since most face-matchers work with gray-scale face images, we used gray-scale images in all experiments to allow for a fair comparison between matchers based on the same input data.} face image $I_{\text{orig}}$, of size $224{\times}224{\times}1$, fused with a face prototype belonging to the same gender ($P_{\text{sm}}$). 
After the fused input image was passed through the encoder and decoder networks, the face prototypes ($P_{\text{sm}}$ prototype face image from the same gender as input image, or $P_{\text{op}}$ the prototype face image of the opposite gender) are added as additional channels to the resulting $128$-channel feature-map representation. Finally, a $1{\times}1$-convolutional operation is used to reduce the number of channels in the resulting feature-maps to a  $224{\times}224{\times}1$-dimensional output image, which is denoted as $I_{\text{sm}}^\prime$ or $I_{\text{op}}^\prime$, depending on the type of prototype used by the decoder:
\begin{equation}
\begin{array}{l}
I_{\text{sm}}^\prime = \text{SAN}(I_{\text{orig}};P_{\text{sm}}), \; \text{and} \\\\
I_{\text{op}}^\prime = \text{SAN}(I_{\text{orig}};P_{\text{op}}).
\end{array}
\end{equation}
These output images, $I_{\text{sm}}^\prime$ and $I_{\text{op}}^\prime$, are then passed to both the auxiliary face matcher and the auxiliary gender classifier. The auxiliary face matcher predicts whether the original and the perturbed face images belong to the same individual via a face match score. The gender classifier predicts the gender of the input and output images via gender probabilities for male and female.\footnote{In this paper, we have assumed binary labels for gender;  however, it  must be noted  that societal and personal interpretation of gender can result in many more classes. }
For the auxiliary face matcher, the pre-trained, publicly available VGG-face model~\cite{parkhi_deep_2015} is used, which computes the face representation vectors for an input face image, and the similarity between two face representation vectors determines the associated match-score.   

Three different loss functions are defined based on the outputs from the autoencoder, the auxiliary gender classifier, and the auxiliary face matcher. The first component of the  loss function, $\mathcal{J}_D$, measures the pixelwise dissimilarity between the input and the output from the same-gender prototype $I_{\text{sm}}^\prime$, which is used to ensure that the autoencoder subnetwork is able to construct realistic face images:  
\begin{eqnarray}
\mathcal{J}_D(I_{\text{orig}},I_{\text{sm}}^\prime) = \frac{1}{h{\times}w} \sum_{i=1}^{h{\times}w} \mathcal{H}(I_{\text{orig}}^{(i)},I_{\text{sm}}^{\prime(i)}),
\end{eqnarray}
where $\mathcal{H}$ indicates the cross-entropy function for the binary case, defined as
\begin{equation}
\mathcal{H}(p,q) = -\left(p \log (q) + (1-p) \log (1-q)\right).
\end{equation}

The second loss term, $\mathcal{J}_M$, is the squared $L^2$ distance between the face representation vectors obtained from the auxiliary face matcher (VGG-face network~\cite{parkhi_deep_2015}) for the input image and the perturbed output, making the autoencoder learn how to perturb face images such that the accuracy of the face matcher is retained:
\begin{equation}
\mathcal{J}_M(I_{\text{orig}},I_{\text{op}}^\prime) = \|\mathcal{R}_{M}(I_{\text{orig}}) - \mathcal{R}_{M}(I_{\text{op}}^\prime)\|_2^2,
\end{equation}
where $\mathcal{R}_{M}(I)$ and $\mathcal{R}_{M}(I_{\text{op}}^\prime)$ indicate the face representation vectors for the input image and the perturbed output based on the opposite-gender prototype.

Finally, the third loss term, $\mathcal{J}_G$, is the cross-entropy loss function applied to the gender probabilities computed by the auxiliary gender classifier, $G$, on the two perturbed output images. Here, the ground-truth label $y$ of the input image is used for $I_{\text{sm}}^\prime$, but the reverse ($1-y$) is used for $I_{\text{op}}^\prime$:
\begin{equation}
\mathcal{J}_G(y,I_{\text{sm}}^\prime,I_{\text{op}}^\prime) = \mathcal{H}(y,G(I_{\text{sm}}^{\prime(k)})) + \mathcal{H}(1-y,G(I_{\text{op}}^{\prime(k)})).
\end{equation}
The total loss, $\mathcal{J}_{tot}$, is the weighted sum of the three individual loss functions described in the previous paragraphs,
\begin{equation}\label{eq:tot-loss}
\mathcal{J}_{tot} = \lambda_1 \mathcal{J}_D + \lambda_2 \mathcal{J}_M + \lambda_3 \mathcal{J}_G,
\end{equation}
where the parameters $\lambda_i$ are the relative weighting terms that can be chosen uniformly or adjusted via hyperparameter optimization. 

In the remaining part of the paper, we use notation $I^\prime$ for the output of a SAN model on a face image $I_\text{orig}$ when using the opposite-gender prototype, i.e.,  $I^\prime=\text{SAN}(I_\text{orig};P_{\text{op}})$.

Based on our previous study~\cite{mirjalili_gender_2018}, we employed a data augmentation and resampling scheme for training the auxiliary gender classifiers as a means to diversify the SAN models.  In particular, by resampling the instances belonging to the underrepresented race in the CelebA~\cite{liu_deep_2015_long} dataset, we aimed to balance the racial distribution in the training data.  In this regard, we generated five resampled training datasets, where in each one a random disjoint subset of samples from the underrepresented race was replicated $40$ times.  This is an effort to enhance the diversity among the SAN models in an ensemble.  The resampling approaches that are used to mitigate the imbalances in the different training datasets employed in this study are described in~\cite{mirjalili_gender_2018}.

\subsection{Training and Evaluation of an Ensemble SAN model}\label{subsec:parallel-training}

\begin{figure}
\begin{center}
   \includegraphics[width=0.8\linewidth]{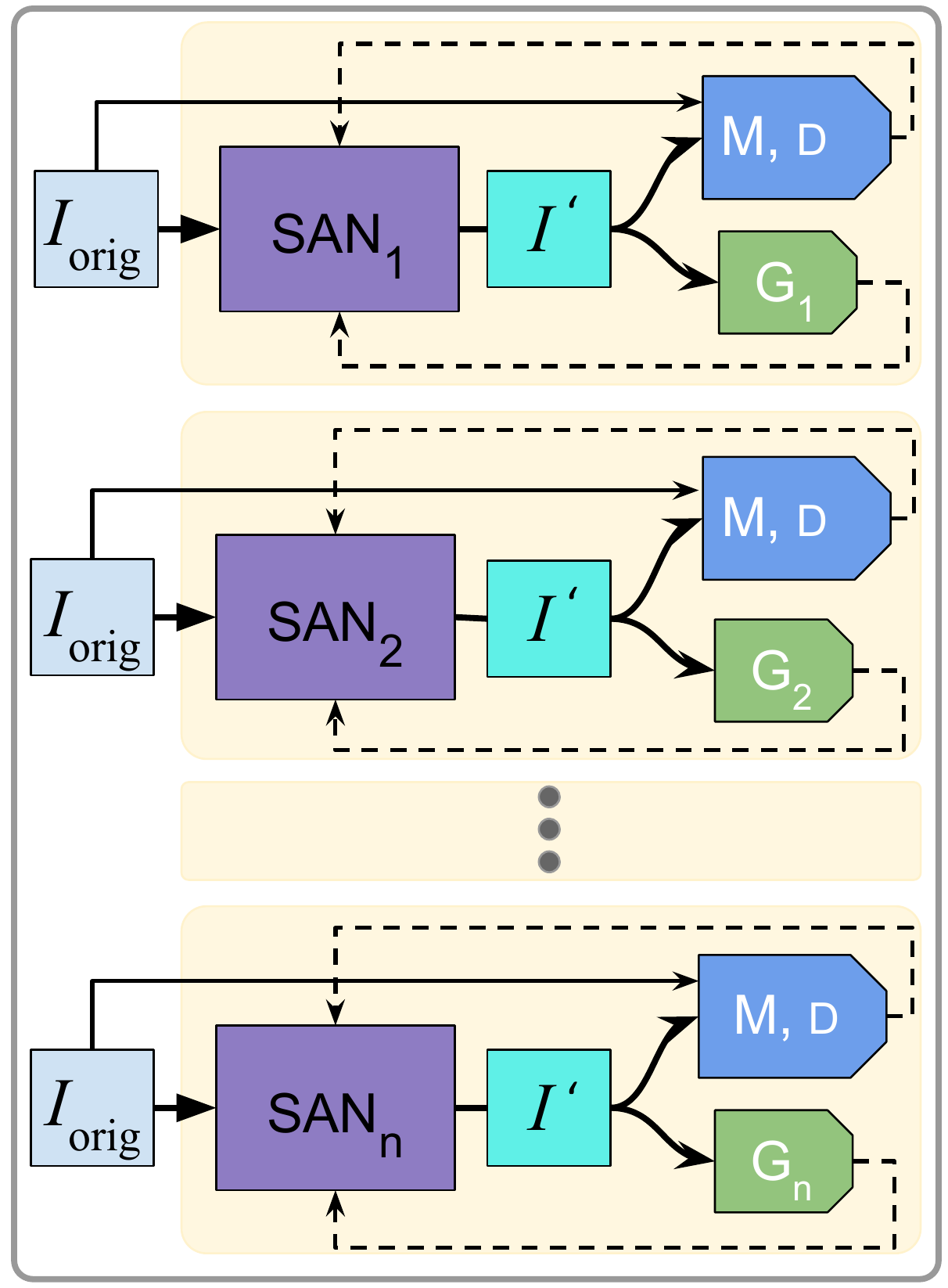}
\end{center}
   \caption{Illustration of an ensemble SAN, where individual SAN models are trained \emph{independent} of each other using $n$ diverse, pre-trained, auxiliary gender classifiers ($\mathcal{G}=\{G_1, G_2, ..., G_n\}$), and a face matcher $M$ that computes face representation vectors for both input face image $I_\text{orig}$ and the output of the SAN model. $D$ refers to a module that computes pixelwise dissimilarity between an input and output face image.}
\label{fig:parallel-ensan-arch}
\end{figure}

In our previous work~\cite{mirjalili_gender_2018}, we proposed an ensemble approach for generalizing SAN models to unseen gender classifiers. The objective of an ensemble SAN was to create $n$ SAN models such that their union can span a larger subset of the hypothesis space compared to a single SAN model. Therefore, for a new test image and an arbitrary unseen gender classifier, $G$, it is likely that at least one of these SAN models in the ensemble is able to confound $G$. For training an ensemble of SANs, we start with $n$ auxiliary gender classifiers, $\mathcal{G}=\{G_1, G_2, ..., G_n\}$, which were trained using different data augmentation schemes (to achieve higher diversity among classifiers), and a pre-trained face matcher $M$. Then, we train $n$ SAN models, where $\text{SAN}_i$ is associated with the auxiliary gender classifier $G_i$, as shown in Fig.~\ref{fig:parallel-ensan-arch}. According to the original SAN model proposed in~\cite{mirjalili_semi_2018}, the loss function for training each model is composed of three components: gender loss, matching loss, and pixelwise dissimilarity loss (Eq.~\ref{eq:tot-loss}). Note that the ensemble of SAN models described with this setting can be trained in parallel since each SAN model is independent of others, and each individual SAN model takes unmodified images as input (Fig.~\ref{fig:parallel-ensan-arch}).  

\begin{figure}
\begin{center}
   \includegraphics[width=0.8\linewidth]{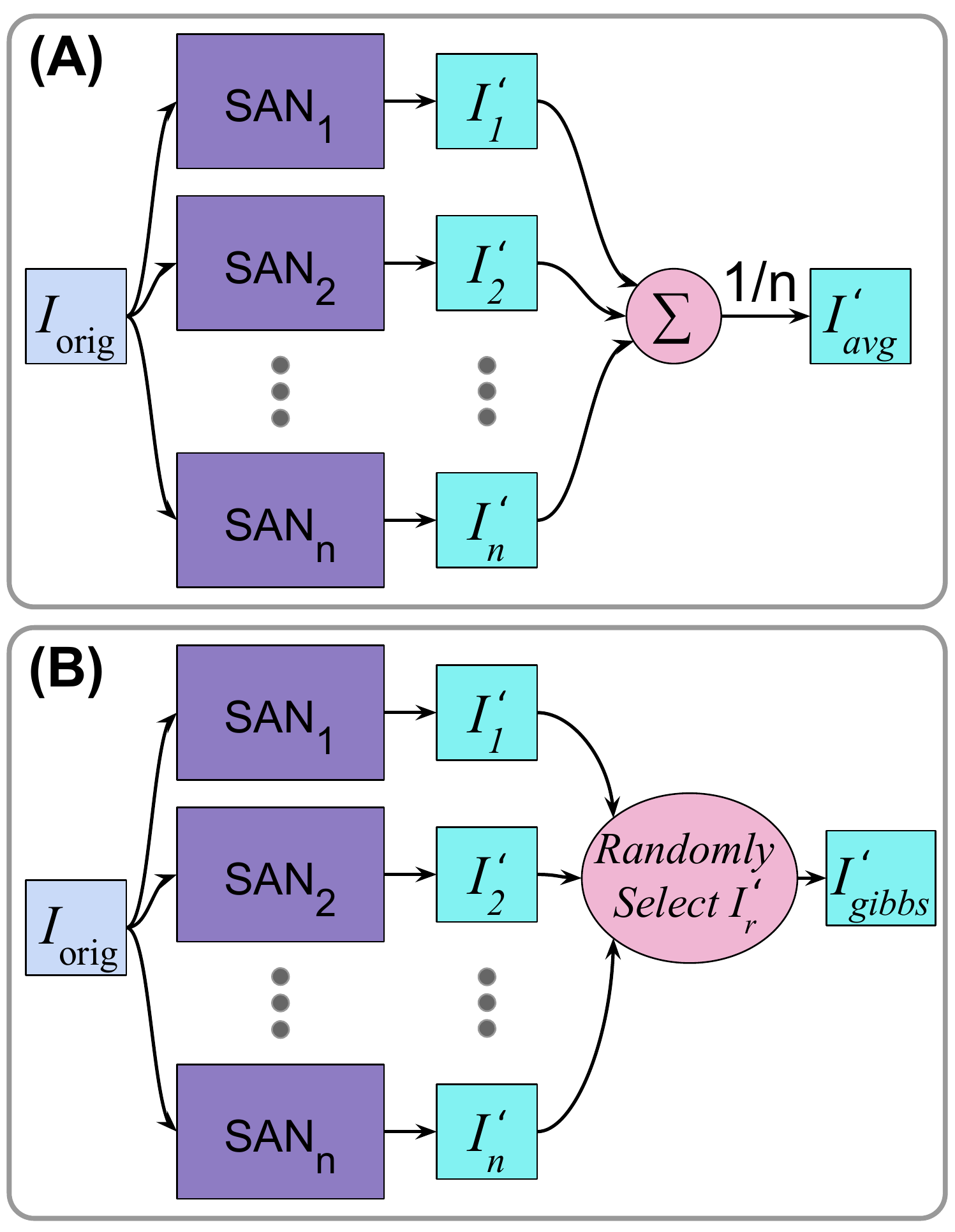}
\end{center}
   \caption{Two approaches for evaluating an ensemble of SAN models: Combining a set of $n$ SAN models trained in the ensemble by (A) averaging $n$ output images, and (B) randomly selecting an output (Gibbs).}
\label{fig:parallel-ensan-eval}
\end{figure}

Evaluation of an ensemble of models, that were trained independently, can be performed in two ways:
\begin{enumerate}
\item Averaging: Evaluating the ensemble of SANs by computing the average output image from the set of $n$ outputs as shown in Fig.~\ref{fig:parallel-ensan-eval}-A.
\item Gibbs: Randomly selecting the output of one SAN model (Fig.~\ref{fig:parallel-ensan-eval}-B).
\end{enumerate}  
These two ensemble-based methods serve as a basis for the comparison with the proposed FlowSAN method, which is described in the following section.

\subsection{FlowSAN: Connecting Multiple SAN Models}\label{subsec:stackeing-sans}

Assume there exists a large set of gender classifiers $\mathcal{G}=\{G_1, G_2, ..., G_g\}$, where each $G_i(I)$ predicts the probability that a face image $I$ belongs to a male individual. Furthermore, suppose there exists a set of $m$ face-matchers denoted by $\mathcal{M}=\{M_1, M_2, ..., M_m\}$, where each $M_i(I_a, I_b)$ computes the match score between a pair of face images, $I_a$ and $I_b$. 
Our goal is to design an {\em ensemble} of $n$ SAN models, $\mathcal{E} = \langle S_1, S_2, ..., S_n\rangle$, that, once they are sequentially stacked together, can be shown to generalize to confound unseen gender classifiers in $\mathcal{G}$. 
We hypothesize that stacking diverse SANs sequentially would have a cumulative effect, where each SAN adds perturbations to an input image that confound a particular gender classifier. Therefore, stacking SANs would enhance their generalizability in terms of decreasing the performance of multiple, diverse gender classifiers. 

We define a recursive function $\Psi_\mathcal{E}(I_{\text{orig}},t)$ for stacking SAN models in $\mathcal{E}=\{\text{SAN}_1, ..., \text{SAN}_n\}$, as follows:
\begin{equation}
\Psi_\mathcal{E}(I_{\text{orig}},t) = \begin{cases}
\text{SAN}_1(I_{\text{orig}}) & \text{if } t=1,\\
\displaystyle \text{SAN}_t\left(\Psi_\mathcal{E}(I_{\text{orig}},t-1)\right) & \text{otherwise.}\\
\end{cases}
\end{equation}

\noindent By varying $t$ from $1$ to $n$, $\Psi_\mathcal{E}(I_{\text{orig}},t)$ produces a sequence of $n$ output images $\langle I_1^\prime, I_2^\prime, ..., I_n^\prime\rangle$:
\begin{itemize}
\item $t=1  \rightarrow\ I_1^\prime = \Psi_{\mathcal{E}}(I_{\text{orig}},1) = \text{SAN}_1(I_{\text{orig}})$,
\item $t=2 \rightarrow\ I_2^\prime = \Psi_{\mathcal{E}}(I_{\text{orig}},2)= \text{SAN}_2\left(\text{SAN}_1(I_{\text{orig}})\right)$,
\item ...
\item $t=n \rightarrow\ I_n^\prime = \Psi_{\mathcal{E}}(I_{\text{orig}},n)=\text{SAN}_n\left(...\ \text{SAN}_1(I_{\text{orig}})\right)$.
\end{itemize}
In particular, we hypothesize that for each $G_i\in \mathcal{G}$, the stacking of SAN models will progressively confound  $G_i$. Since the individual SAN models were trained to have a minimal impact on face matching performance, we further hypothesize that the perturbations introduced in the output face images $\langle I_1^\prime, ..., I_n^\prime\rangle$ from the stacked SAN models should not substantially affect the face recognition performance of the matchers in $\mathcal{M}$.

\vspace{6px}
\noindent\textbf{Training Procedure for the FlowSAN Model}
\vspace{2px}

\noindent The goal of this work is to develop a model that leverages the image perturbations induced by individual, diverse SAN models to broaden the spectrum of diverse gender classifiers that can successfully be confounded. To accomplish this goal, we designed and evaluated the FlowSAN model, where multiple individually-trained SAN models were sequentially combined.

\begin{figure*}[h]
\begin{center}
   \includegraphics[width=0.98\linewidth]{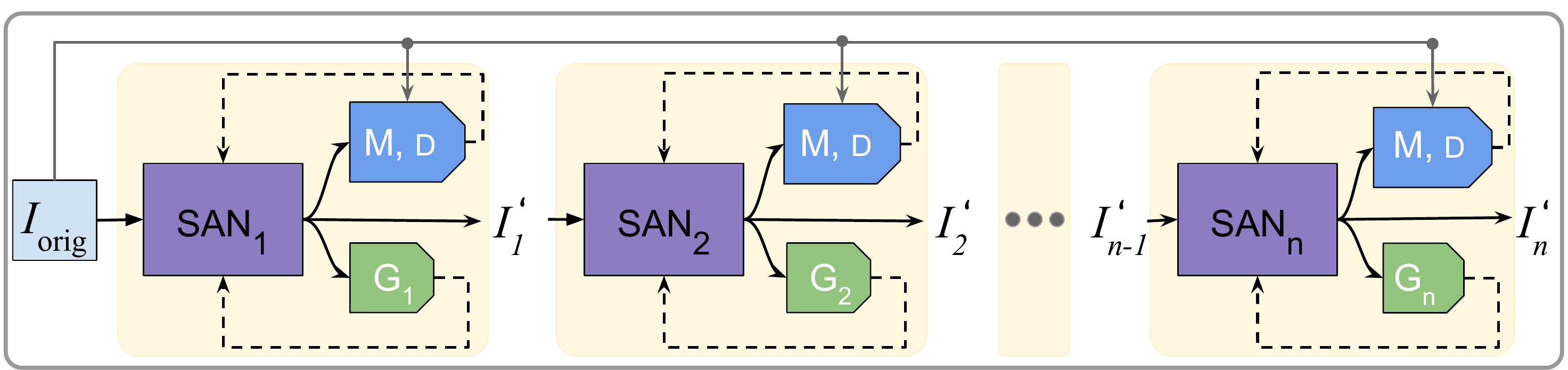}
\end{center}
   \caption{An illustration of a FlowSAN model: $n$ SAN models are trained \emph{sequentially} using $n$ auxiliary gender classifiers ($\mathcal{G}=\{G_1, G_2, ..., G_n\}$), and a face matcher $M$ that computes face representation vectors for both input image $I$ and the output of SAN model. Both auxiliary face matcher and the dissimilarity unit (D) use the original image along with the output of their corresponding SAN.}
\label{fig:sequential-ensan-arch}
\end{figure*}

This section describes the training procedure for the FlowSAN model, where SAN models $i=1, ..., n$ are trained in sequential order, each with their corresponding auxiliary gender classifier and an auxiliary face matcher, which is common among all SANs. The first SAN model, $\text{SAN}_1 \in \mathcal{E} = \{\text{SAN}_1, ..., \text{SAN}_n\}$, takes the original image as input and generates a perturbed output, $I_1^{\prime}$, while using the auxiliary gender classifier $G_1$ during its training. Then, once $\text{SAN}_1$ is trained, the entire training dataset is transformed by $\text{SAN}_1$, and the transformed data is then used for training the next SAN model while using its corresponding auxiliary gender classifier.  This process is repeated  for SAN models $i=1, ..., n$, to obtain $n$ SAN models that are trained in sequential order.  Note that the matching loss is computed between face representation vectors (generated by a face matcher) of the SAN output with that of the corresponding original face image, as opposed to the input to the SAN model (which is already perturbed for $i\ge 2$).  This is to ensure that the matching performance does not substantially decline as the sequence is expanded.  Furthermore, we considered three different scenarios for the pixelwise dissimilarity loss:
\begin{enumerate}
\item Omitting the pixelwise dissimilarity loss term;
\item pixelwise dissimilarity with respect to the input, i.e., $I_{i-1}^\prime$ for $\text{SAN}_{i}$;
\item pixelwise dissimilarity loss with respect to the original image $I_{orig}$ for each of SAN models $i=1, ..., n$.
\end{enumerate}
 We evaluated all three different pixelwise loss function schemes listed above. However, we were unable to observe any noticeable differences except for some cases where the third scheme slightly outperformed the other two. Therefore, we only report the results of the third case in this paper. The training procedure is illustrated in Fig.~\ref{fig:sequential-ensan-arch}.

\vspace{6px}
\noindent\textbf{Evaluating the FlowSAN Model}
\vspace{2px}

\noindent During the model evaluation, the auxiliary networks (the auxiliary gender classifiers and auxiliary face matchers) from the individual SANs are discarded, and the $n$ SAN models are stacked in the same sequence they were trained, in order to enhance their generalizability to arbitrary gender classifiers.  
In the FlowSAN model, the first SAN model ($\text{SAN}_1$) takes an original image ($I_{\text{orig}}$) as input and generates a perturbed output image $I_1^\prime$. This output image is then passed into the next SAN model in the sequence to obtain $I_2^\prime$, and so forth. In general, the $i$th SAN model ($\text{SAN}_i$ for $i=2,...,n$) takes the output of the previous SAN model ($I_{i-1}^\prime$) as input and generates the perturbed output $I_i^\prime$.


\section{Experiments and Results}

We designed two different protocols for training $n$ SAN models:
\begin{description}
\item[(a)] Training an ensemble of SANs independent of each other as described in~\cite{mirjalili_gender_2018} (see Section~\ref{subsec:parallel-training});
\item[(b)] Training the FlowSAN model using the sequential  procedure described in Section~\ref{subsec:stackeing-sans}. 
\end{description}

Protocol (a) was adapted from~\cite{mirjalili_gender_2018} and is further described in Section~\ref{subsec:parallel-training}.
For evaluating models trained in the ensemble, we applied two techniques: 1) taking the average output from SAN models which we denote as Ens-Avg, and 2) randomly selecting the output which we denote as Ens-Gibbs. 
In addition, similar to~\cite{mirjalili_gender_2018}, we also define the {\em oracle best-perturbed} sample for a specific gender classifier, $G$:
\begin{equation}
\text{best}(I;\mathcal{E},G) = \begin{cases}
\displaystyle \argmin_{\text{SAN}_i \in \mathcal{E}}G(\text{SAN}_i(I)) & \text{if }
y=1,\\
\displaystyle \argmax_{\text{SAN}_i \in \mathcal{E}}G(\text{SAN}_i(I)), & \text{otherwise}.\end{cases}
\end{equation}
The results of best-perturbed samples are denoted as Ens-Best. This analysis indicates which output from the ensemble model $\mathcal{E}$ has resulted in the highest prediction error for a particular gender classifier $G$ if the best output is selected.

The training of the FlowSAN model was initiated from the pre-trained individual SAN models in~\cite{mirjalili_gender_2018} and then trained for $10$ additional epochs on the CelebA-train subset~\cite{liu_deep_2015_long} (see Table~\ref{tab:datasets}) using the training procedure described in Section~\ref{subsec:stackeing-sans}.  Then, the models were stacked successively to generate a sequence of perturbed output images, $\langle I_1^\prime, \dots, I_n^\prime\rangle$.

As the FlowSAN model conceals the gender information in face images incrementally, it naturally produces a sequence of perturbed face images, where the length of this sequence is determined by its ensemble size. By varying the size of the ensemble, we can have a fair comparison between the ensemble approach vs. the FlowSAN model, such that the number of SANs used to obtain an output from the ensemble model is consistent with the number of SANs that are used to generate the output from the FlowSAN model.

For model evaluation and comparison, we used four test datasets: CelebA-test~\cite{liu_deep_2015_long}, MORPH-test~\cite{ricanek_morph_2006}, MUCT~\cite{milborrow_muct_2010_long}, and RaFD~\cite{langner_presentation_2010}. The number of male and female individuals in each dataset is listed in Table~\ref{tab:datasets}.

\begin{table}
\begin{center}
\caption{Overview of datasets used in this study. The letters in the ``Usage'' column indicate the tasks for which the datasets were used. a: training auxiliary gender classifiers, b: SAN training, c: SAN evaluation, d: constructing unseen gender classifiers used for evaluating SAN models.}
\label{tab:datasets}
\begin{threeparttable}
\centering
\small
\begin{tabular}{lccc}
 \toprule
 {\bf Dataset}  & {\bf \#male}& {\bf \#female}&  {\bf Usage} \\
\midrule 
CelebA-train & {73,549} &  {103,772} & a, b\\ 
CelebA-test & {7,929} & {11,511}  & c\\ 
MORPH-train &  {41,587} &  {7,567}  & d\\ 
MORPH-test & {4,643} & 863 & c\\ 
LFW & {10,064} & {2,905}  & d\\ 
MUCT & {1,844} & {1,910}   & c\\ 
RaFD & {1,008} & 600 & c\\ 
\bottomrule 
\end{tabular} 
\end{threeparttable}
\end{center}
\end{table}

\subsection{Performance in Confounding Unseen Gender Classifiers}

In order to evaluate the generalization performance of the three ensemble-based methods discussed in the previous section (Ens-Avg, Ens-Gibbs, Ens-Best) as well as the proposed FlowSAN model, we considered six independent gender classifiers. The experiments designed in this section assess how well the proposed models are able to confound gender classifiers that were unseen during training. These six gender classifiers include three models that were already trained: a commercial-of-the-shelf gender classifier (G-COTS), IntraFace~\cite{de_la_torre_intraface_2015_long}, AFFACT~\cite{gunther_affact_2017}, and three CNN models built in-house, which we refer to as CNN-1, CNN-2 (trained using MORPH-train and LFW, respectively), and CNN-3 (trained on the union of MORPH-train and LFW). Note that these three CNN models have shown a similar level of performance on the original test-sets, compared to the other three pre-trained gender predictors.

\begin{figure*}
\begin{center}
   \includegraphics[width=0.95\linewidth]{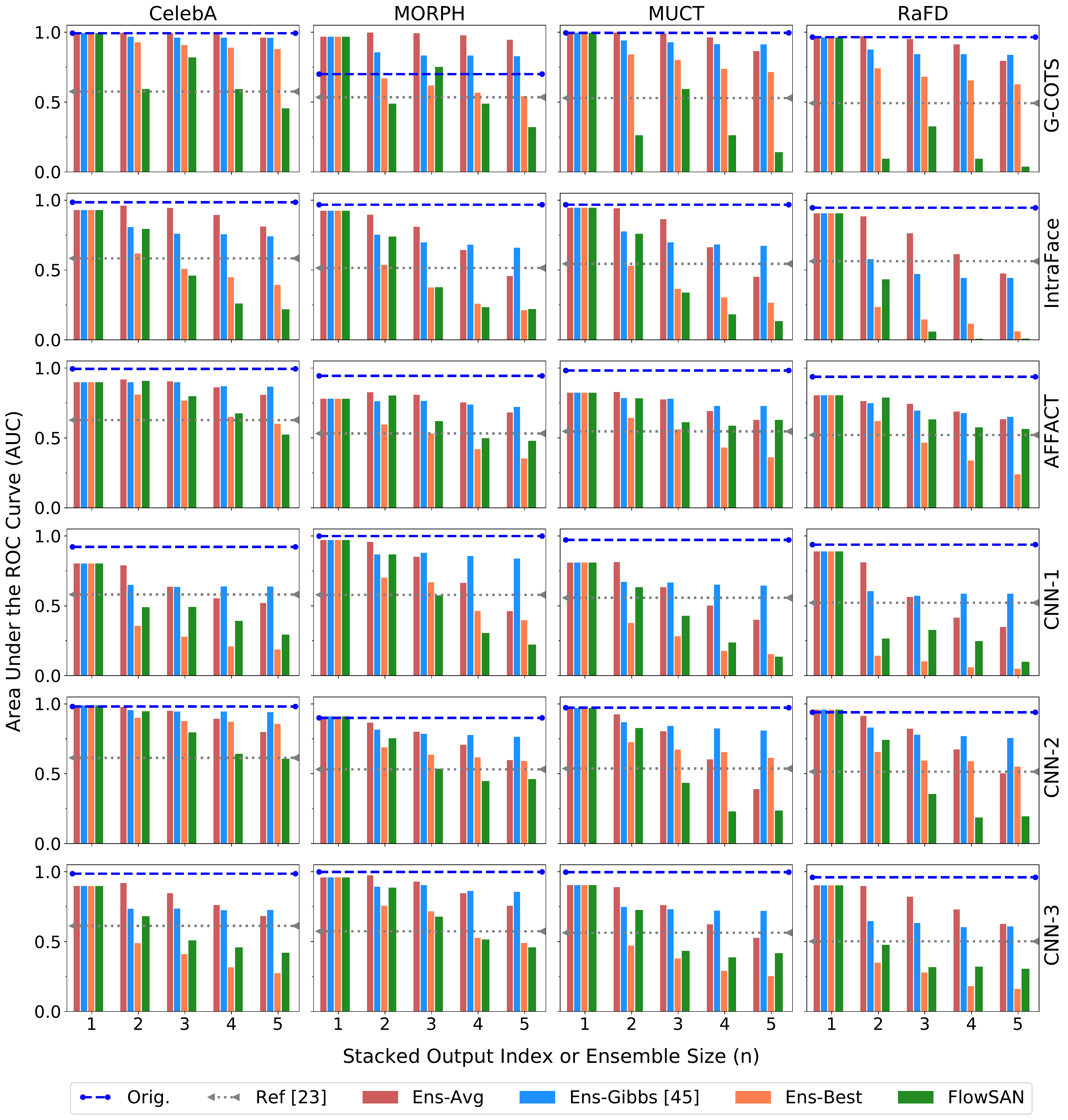}
\end{center}
   \caption{Area under the ROC curve (AUC) measured for the six unseen gender classifiers (CNN-3, CNN-2, CNN-1, AFFACT, IntraFace, and G-COTS) on the test partitions of the four different datasets (CelebA, MORPH, MUCT, and RaFD). The gender classification performance on the original images ("Orig.") is shown (blue dashed line) as well as the perturbed samples using the three ensemble-based models (Ens-Avg, Ens-Gibbs, Ens-Best) the proposed FlowSAN model, and the face mixing approach~\cite{othman_privacy_2014} (gray dashed line). The index (1, 2, ..., 5) on the x-axis indicates the sequence of outputs $\langle I_1^\prime, I_2^\prime, ..., I_5^\prime\rangle$ obtained by varying the ensemble size, $n$.  In almost all cases, stacking three SAN models results in an AUC of approximately 0.5 (a perfectly random gender prediction).}
\label{fig:unseen-gpreds}
\end{figure*}

Fig.~\ref{fig:unseen-gpreds} shows the area under the ROC curve as a performance metric for evaluating the generalization performance of each unseen gender classifier on the four independent test datasets.  
The performance of these gender classifiers on the original images (before perturbations), as well as the outputs from the mixing approach by~\cite{othman_privacy_2014}, is also shown for comparison.

In all cases, the FlowSAN approach results in lower AUC values (lower is better) of predictions made by unseen gender classifiers (Fig.~\ref{fig:unseen-gpreds}) compared to the ensemble models Ens-Avg and Ens-Gibbs. In fact, the results of the stacking SAN models are almost on par with the oracle best-perturbed samples (Ens-Best) for each gender classifier. In some cases, the FlowSAN model even outperforms Ens-Best. {\bf It is important to note that selecting the best-perturbed sample (from the individual SAN models) for each gender classifier without {\em a priori} knowledge of the classifier is infeasible in practice. Yet, we are able to outperform the best result using the FlowSAN model in several cases.}

Note that in a real privacy application, reaching a near random gender prediction performance ($\text{AUC} \approx 0.5$, and Equal Error Rate (EER) $\approx 0.5$) is desired for gender anonymization. As it can be seen in Fig.~\ref{fig:unseen-gpreds}, both Ens-Avg and Ens-Gibbs methods produce samples that are mostly incapable of lowering the  AUC of the unseen gender classifiers below $0.75~\text{AUC}$. 
Based on the results shown in Fig.~\ref{fig:unseen-gpreds} (and the EER results shown in Fig.~\ref{supfig:unseen-gpreds-eer}), it is evident that, in the majority of cases, a sequential stacking of three SAN models via FlowSAN produces the desired behavior in terms of face gender-anonymization, i.e.,   $\text{AUC} \approx 0.5$ (similarly, $\text{EER} \approx 0.5$).
Although, in some cases, the \nth{5} output from Ens-Avg and Ens-Gibbs resulted in a low, desired  AUC of $\approx 0.5$, it also has a substantially detrimental effect on the face matching performance, as discussed in Section~\ref{sec:unseen-matchers}. 

As a result, we conclude that stacking three SAN models in FlowSAN is sufficient to achieve the best gender label anonymization performance across a set of different, unseen gender classifiers and face image datasets. Stacking fewer than three models affects unseen gender classifiers substantially less, and stacking more than three models induces such strong perturbations that flipping the predicted labels could again de-anonymize the perturbed face images with respect to their gender labels.

We shall note that our study was not the first to confound gender classifiers to produce random predictions. In ~\cite{othman_privacy_2014}, researchers proposed a face mixing approach that also leads to successful gender anonymization  (approximately $0.5$ AUC gender prediction performance for a specific gender classifier); however, this approach was unable to retain the face matching utility. In different studies, the researchers were able to retain face matching utility but without generalizing to arbitrary gender classifiers~\cite{mirjalili_soft_2017,chhabra_anonymizing_2018}. Thus, the FlowSAN model we propose in this paper presents the first successful approach for satisfying both objectives: concealing gender information and retaining matching performance to a satisfactory degree across a variety of independent gender classifiers and face matchers.

\begin{figure*}
\begin{center}
   \includegraphics[width=0.95\linewidth]{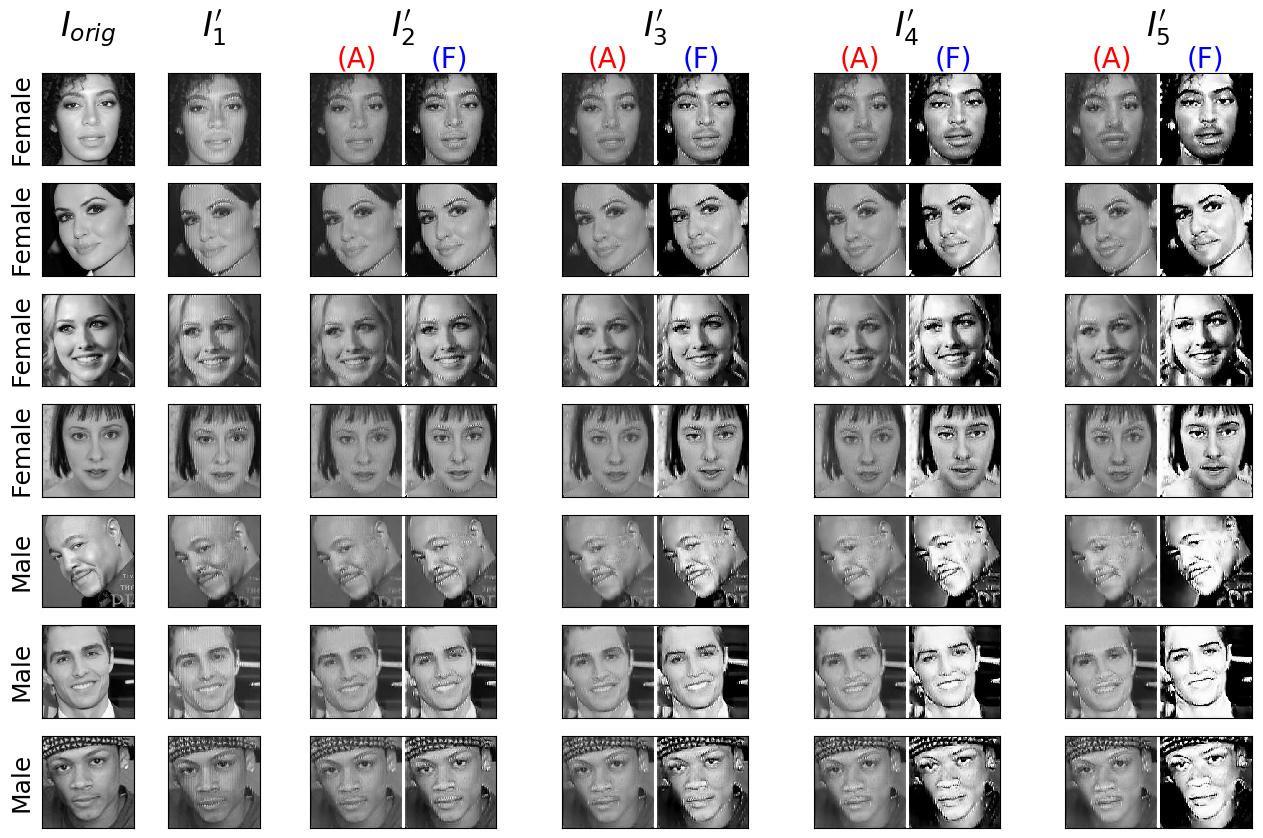}
\end{center}
   \caption{A randomly selected set of examples showing input face images and their outputs from $I_1^\prime$ to $I_5^\prime$ using (A) the ensemble model, Ens-Avg, and (F) using the FlowSAN model.}
\label{fig:examples}
\end{figure*}

\subsection{Retaining the Performance of Unseen Face Matchers}
\label{sec:unseen-matchers}

To assess the effect of the gender perturbations on the matching accuracy, we considered four different unseen face matchers. This includes a commercial-of-the-shelf face matcher (M-COTS), which has shown state-of-the-art performance in face recognition, as well as three publicly available algorithms that provide face representation vectors: DR-GAN~\cite{tran_disentangled_2017}, FaceNet~\cite{schroff_facenet_2015}, and OpenFace~\cite{amos_openface_2016}.  For the latter three models, we measured the cosine similarity between face representation vectors obtained from the original images and face representation vectors obtained from the SAN-perturbed output images.

\begin{figure*}
\begin{center}
   \includegraphics[width=0.95\linewidth]{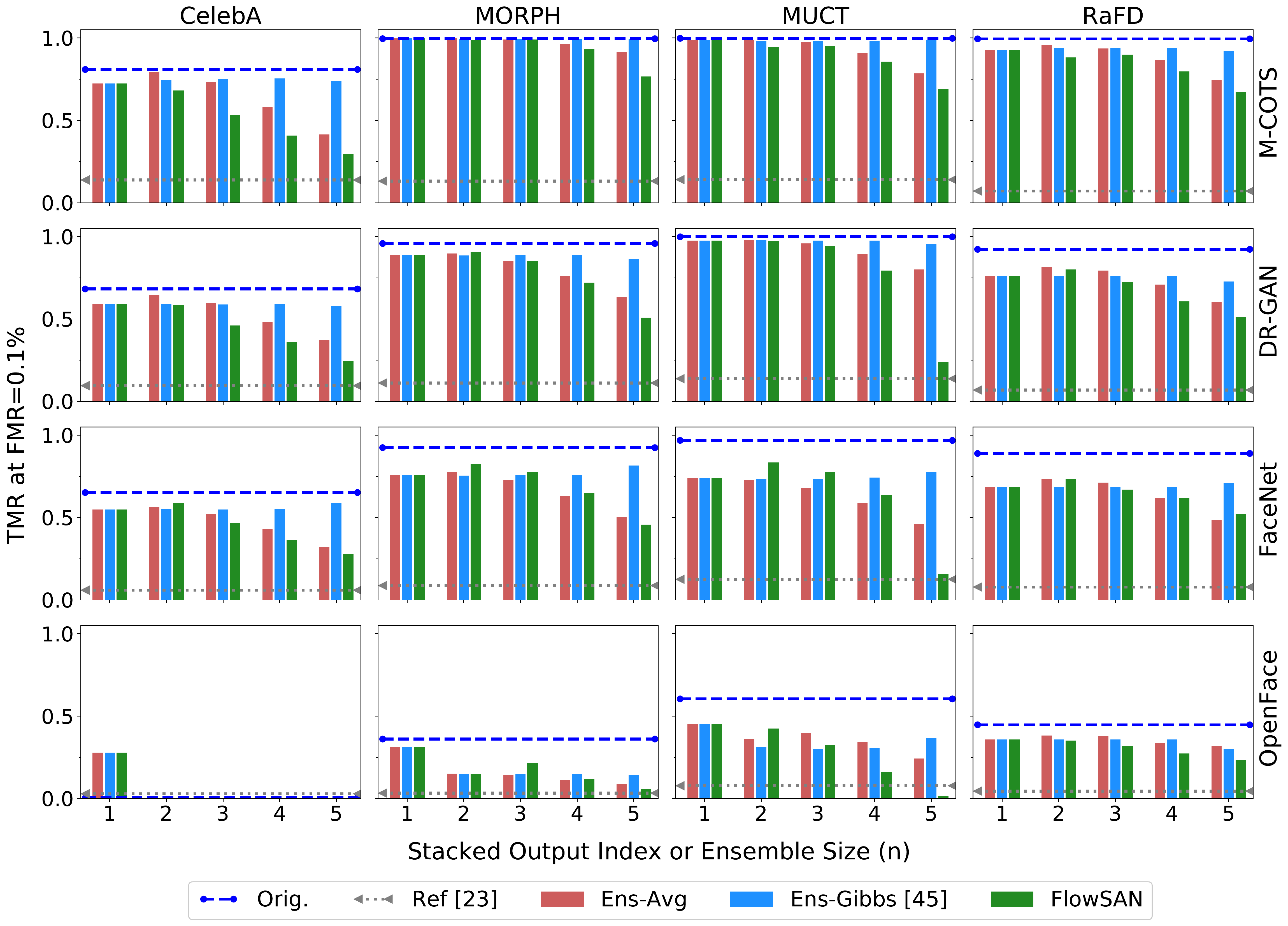}
\end{center}
   \caption{True Match Rate (TMR) values at False Match Rate (FMR) of $0.1\%$ obtained from four unseen face matchers, M-COTS, DR-GAN, FaceNet, and OpenFace on the original images as well as perturbed outputs after applying stacking SAN models and the ensemble models (Ens-Avg and Ens-Gibbs). Note that the matchers\textquotesingle  ~performance obtained after applying the first three SANs in the FlowSAN model is close to the original performance, but it further diminishes when the sequence is extended.}
\label{fig:unseen-matchers}
\end{figure*}

Fig.~\ref{fig:unseen-matchers} shows the True Match Rate (TMR) values at False Match Rate (FMR) of $0.1\%$ for different ensemble methods.  In most cases, the performance of the face matchers regarding the first three outputs ($I_1^\prime$, $I_2^\prime$, and $I_3^\prime$) is similar and relatively close to the matching performance on original images.  We note that stacking three SANs in FlowSAN yields the desired performance with regard to confounding unseen gender classifiers. Therefore, the evaluation of the face matching performance for stacking more than three SANs $I_3^\prime$ (i.e., $I_4^\prime$ and $I_5^\prime$) is only included for completeness. 

Comparing the performance of face matchers for equal values of $n$, we observe that the face matchers appear to perform slightly better on outputs produced by the ensemble model compared to the FlowSAN model. However, the extent to which the gender classification performance is reduced by the two models is not the same for equal values of $n$ (Table~\ref{tab:overall-performance}). The ensemble model requires at least $n=5$ individual SAN models to be able to confound unseen gender classifiers to reach the same level of gender anonymization as the FlowSAN model with $n=3$. Therefore, if we compare the ensemble models with $n=5$ to the FlowSAN model with $n=3$, the face matchers perform substantially better on the face image outputs by the FlowSAN model (Fig.~\ref{fig:unseen-matchers}).
Further, note that the performance of M-COTS on CelebA on the original images is already as low as $85.6\%$. In fact, all matchers perform poorly on the CelebA dataset, which may be due to different face orientations captured in the wild.

\vspace{6px}
\noindent \textbf{Preserving Privacy}
\vspace{2px}

\noindent The overall average performance considering the two target objectives of this study, i.e., confounding gender classifiers and retaining the matching utility of face images, is provided in Table~\ref{tab:overall-performance}. In this analysis, the average EER results of all six gender classifiers over all four evaluation datasets were computed for original images, outputs from Ref.~\cite{othman_privacy_2014}, as well as outputs from the stacking and the ensemble models using $n=3$ and $n=5$. The results clearly show that the  FlowSAN model outperforms the ensemble-based methods, including the oracle-best results.  On the other hand, the average true matching rate (TMR) values, at a false matching rate (FMR) of 0.1\%, are also computed similarly, and the results indicate that the Ens-Gibbs method has the highest performance for both ensemble sizes, while the performance of the FlowSAN model at $n=3$ is ranked as second, but it is very close to that of Ens-Gibbs. The detailed EER results for each gender classifier is provided in Table~\ref{suptab:overall-performance-full}.

\begin{table}
\caption{Comparing the overall average performance of six unseen gender classifiers and four unseen face matchers over the four evaluation datasets using $n=3$ or $n=5$ SAN models. This shows that stacking $3$ SAN models results in gender anonymization $\text{EER} \approx 0.5$, while the the average matching performance is still comparable to the unmodified images as well as the matching performance on the outputs form other existing methods.}
\label{tab:overall-performance}
\begin{center}
\begin{threeparttable}
\small
\begin{tabular}{l|cccc}
 & \multicolumn{2}{c}{\bf Gender:} & \multicolumn{2}{c}{\bf Matching:} \\
 & \multicolumn{2}{c}{\multirow{2}{*}{\bf EER}} & \multicolumn{2}{c}{\bf TMR at}\\ 
 & & & \multicolumn{2}{c}{\bf FMR=0.1\%} \\
 \toprule
Orig. & \multicolumn{2}{c}{10\%} & \multicolumn{2}{c}{76.3\%} \\
Ref~\cite{othman_privacy_2014} & \multicolumn{2}{c}{46\%} & \multicolumn{2}{c}{9.1\%} \\ \hline
 & $n=3$ & $n=5$ & $n=3$ & $n=5$ \\ \cline{2-5} 
Ens-Avg & 23\% & 40\% & 64.9\% & 48.1\%\\
Ens-Gibbs & 29\% & 31\% & 65.2\% & 65.6\%\\
Ens-Best & 48\% & 57\% & -- & -- \\ 
FlowSAN & {\bf 49}\%& {\bf 64}\% & 61.9\% & 35.4\% \\\bottomrule
\end{tabular}
\end{threeparttable}
\end{center}
\end{table}

\vspace{6px}
\noindent \textbf{Computational Efficiency}
\vspace{2px}

\noindent The overall computational cost for training the ensemble-based approach and the FlowSAN model is similar, except that FlowSAN requires an additional data transformation step between each consecutive SAN training. However, the ensemble approach comes with a bigger advantage that the individual SAN models can be trained in parallel, while the SAN models in the FlowSAN model have to be trained sequentially. 

\section{Conclusion}

In this work, we address one of the main limitations of previous gender privacy methods, namely, their inability to generalize across multiple previously unseen gender classifiers. In this regard, we propose the FlowSAN method that sequentially combines diverse perturbations for an input face image to confound the gender information with respect to an arbitrary gender classifier. We compared the performance of the proposed FlowSAN model with two ensemble-based approaches: 1) using the average output of SAN models trained independent of each other (Ens-Avg); 2) randomly selecting the output from the SAN models in the ensemble (Ens-Gibbs).  

Our experiments show that the FlowSAN method outperforms the other ensemble-based approaches in terms of confounding gender attribute for a range of gender classifiers.
More importantly, while gender classification is successfully confounded, face matching accuracy is retained for all perturbed output face images, thereby preserving the biometric utility of the gender-anonymous face images. 

While this work only focused on confounding gender labels to demonstrate this method's efficacy in hiding soft-biometric attributes, our method can be readily extended and generalized to incorporate other soft-biometric attributes (for example, age and ethnicity), which is subject of future studies.



{
\bibliographystyle{IEEEtran}
\balance
\bibliography{egbib}
}

\section{Supplementary Materials}

\setcounter{figure}{0}
\setcounter{table}{0}
\renewcommand{\thefigure}{S\arabic{figure}}
\renewcommand{\thetable}{S\arabic{table}}

\begin{figure*}
\begin{center}
   \includegraphics[width=0.95\linewidth]{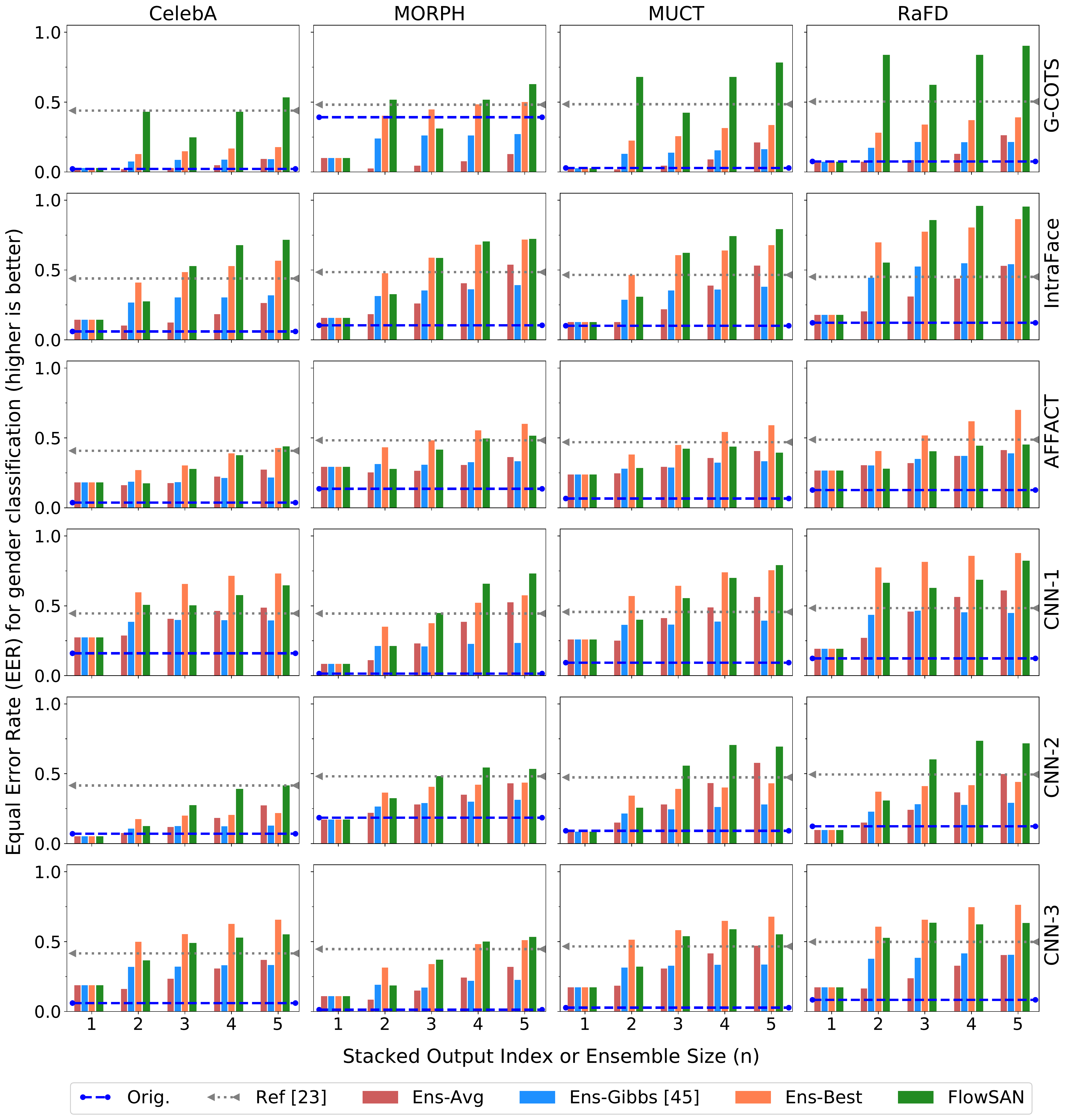}
\end{center}
   \caption{Equal Error Rate (EER) measured for the six unseen gender classifiers (CNN-3, CNN-2, CNN-1, AFFACT, IntraFace, and G-COTS) on the test partitions of the four different datasets (CelebA, MORPH, MUCT, and RaFD). The gender classification performance on the original images ("Orig.") is shown (blue dashed line) as well as the perturbed samples using the three ensemble models (Ens-Avg, Ens-Gibbs, Ens-Best), the proposed FlowSAN model, and the face mixing approach~\cite{othman_privacy_2014} (gray dashed line). The index (1, 2, ..., 5) on the x-axis indicates the sequence of outputs $\langle I_1^\prime, I_2^\prime, ..., I_5^\prime\rangle$ obtained by varying the ensemble size, $n$.}
\label{supfig:unseen-gpreds-eer}
\end{figure*}

\begin{table*}
\caption{Comparing the overall average Equal Error Rate (EER) of six unseen gender classifiers averaged over all four evaluation datasets (CelebA-test, MORPH-test, MUCT, and RaFD), higher is better. Note that the Ens-Best method is the result of ``oracle best'' selected classifier from an ensemble of multiple SANs, which assumes knowledge of the gender classifier. While this is impractical in a real-world privacy application, we show the results for comparison purposes.}
\label{suptab:overall-performance-full}
\begin{center}
\begin{threeparttable}
\small
\begin{tabular}{l | cc | cccc | cccc} \toprule 
 Gender & \multirow{2}{*}{Orig.} & \multirow{2}{*}{Ref.~\cite{othman_privacy_2014}}& \multicolumn{4}{c|}{$n=3$} & \multicolumn{4}{c}{$n=5$} \\
 Classifier & & & Ens-Avg & Ens-Gibbs & Ens-Best & FlowSAN & Ens-Avg & Ens-Gibbs & Ens-Best & FlowSAN \\ \hline
  G-COTS & 0.13 & 0.48 & 0.05 & 0.18 & 0.30 & {\bf 0.40} & 0.17 & 0.18 & 0.35 & {\bf 0.71} \\  
IntraFace & 0.10 & 0.46 & 0.23 & 0.38 & 0.61 & {\bf 0.65} & 0.47 & 0.41 & 0.71 & {\bf 0.80} \\  
AFFACT & 0.09 & 0.46 & 0.26 & 0.28 & {\bf 0.44} & 0.38 & 0.36 & 0.32 & {\bf 0.58} & 0.45 \\  
CNN-1 & 0.10 & 0.46 & 0.38 & 0.36 & {\bf 0.62} & 0.53 & 0.55 & 0.38 & 0.74 & {\bf 0.75} \\  
CNN-2 & 0.12 & 0.47 & 0.23 & 0.23 & 0.35 & {\bf 0.48} & 0.45 & 0.25 & 0.38 & {\bf 0.59} \\  
CNN-3 & 0.05 & 0.46 & 0.23 & 0.30 & {\bf 0.53} & 0.51 & 0.39 & 0.32 & {\bf 0.65} & 0.57 \\ \cline{1-11}
{\bf Average} & 0.10  & 0.46 & 0.23  & 0.29  & 0.48  & {\bf 0.49} & 0.40  & 0.31  & 0.57  & {\bf 0.64}\\ \bottomrule
\end{tabular}
\end{threeparttable}
\end{center}
\end{table*}

\ifCLASSOPTIONcaptionsoff
  \newpage
\fi

\end{document}